\documentclass[lettersize,journal]{IEEEtran}
\usepackage{amsmath,amsfonts}
\usepackage{algorithmic}
\usepackage{algorithm}
\usepackage{array}
\usepackage[caption=false,font=normalsize,labelfont=rm,textfont=rm]{subfig}
\usepackage{textcomp}
\usepackage{stfloats}
\usepackage{gensymb}
\usepackage{url}
\usepackage{verbatim}
\usepackage{graphicx}
\usepackage{cite}
\usepackage{svg}
\usepackage{booktabs}
\usepackage{epstopdf}
\usepackage{xcolor}
\usepackage{multirow} 
\usepackage{tabularx}
\usepackage{makecell}
\usepackage{diagbox}

\begin{document}

\title{Motion Adaptation Across Users and Tasks for Exoskeletons via Meta-Learning}

\author{Muyuan Ma, Xiuze Xia, Lijun Han, Houcheng Li and Long Cheng,~\IEEEmembership{Fellow,~IEEE}}

\markboth{IEEE ROBOTICS AND AUTOMATION LETTERS,~Vol.~XX, No.~XX, JUNE~2025}%
{Ma \MakeLowercase{\textit{et al.}}: Motion Adaptation Across Users and Tasks for Exoskeletons via Meta-Learning}


\maketitle
\pagestyle{empty}  
\thispagestyle{empty} 

\begin{abstract}
Wearable exoskeletons can augment human strength and reduce muscle fatigue during specific tasks. However, developing personalized and task-generalizable assistance algorithms remains a critical challenge. To address this, a meta-imitation learning approach is proposed. This approach leverages a task-specific neural network to predict human elbow joint movements, enabling effective assistance while enhancing generalization to new scenarios. To accelerate data collection, full-body keypoint motions are extracted from publicly available RGB video and motion-capture datasets across multiple tasks, and subsequently retargeted in simulation. Elbow flexion trajectories generated in simulation are then used to train the task-specific neural network within the model-agnostic meta-learning (MAML) framework, which allows the network to rapidly adapt to novel tasks and unseen users with only a few gradient updates. The adapted network outputs personalized references tracked by a gravity-compensated PD controller to ensure stable assistance. Experimental results demonstrate that the exoskeleton significantly reduces both muscle activation and metabolic cost for new users performing untrained tasks, compared to performing without exoskeleton assistance. These findings suggest that the proposed framework effectively improves task generalization and user adaptability for wearable exoskeleton systems.
\end{abstract}


\section{Introduction}
\IEEEPARstart{E}{xoskeletons} are increasingly deployed in industrial, military, and rehabilitation domains to enhance human performance during task-specific operations\cite{ref1}. These exoskeletons are broadly categorized into passive and active exoskeletons. Passive exoskeletons rely on mechanical elements such as springs or dampers to redistribute loads and reduce muscle strain, without the use of powered actuators \cite{ref2}. In contrast, active exoskeletons are equipped with actuators that actively generate assistive torques \cite{ref3}. To provide effective assistance, active exoskeletons need to predict users' joint motion trajectories and generate joint torques \cite{ref4}. This predictive capability allows active exoskeletons to deliver more precise, adaptive, and task-specific support than passive systems.

In the literature, various studies have focused on predicting users' motion trajectories for active exoskeletons to provide assistive support. Since neural activity precedes actual movement \cite{ref5}, a common approach for predicting users' motion intention is to decode neural signals, such as electromyography (EMG) \cite{ref6} and electroencephalography (EEG) \cite{ref7}. \cite{ref7-1} proposed a hierarchical optimization approach for a personalized musculoskeletal model, which uses EMG signals to predict both single-joint and coordinated multi-joint movements. Furthermore, \cite{ref7-2} designed a joint motion predictor that integrates convolutional neural networks with Mamba, enabling simultaneous estimation of joint angles and torques across multiple joints using EMG inputs. \cite{ref9} proposed a novel multimodal architecture, in which subjects controlled the exoskeleton via a brain–computer interface to provide active assistance for reaching tasks in real-world scenarios. To further improve control accuracy, \cite{ref9-1} developed a personalized musculoskeletal model driven by EMG signals. However, biosignals such as EMG and EEG are sensitive to electrode placement and suffer from low signal-to-noise ratios. Consequently, recent methods have explored  predicting users' motion intention based on the combined motion states of the human and the robot, without relying on biosignals \cite{ref9-2}. As presented in \cite{ref10}, an online learning and prediction algorithm based on the dynamic movement primitives (DMP) model was used to estimate human joint trajectories and torques, facilitating assistive control for human walking. In addition, an adaptive DMP-based trajectory prediction method was developed in \cite{ref11} to enable wearable robotic assistance for various discrete lifting movements. In \cite{ref12}, a deep Gaussian process-based method was employed to predict the operator’s gait, enabling real-time estimation of motion intention. The aforementioned methods have shown promising performance in homogeneous tasks and with trained users. However, they lack the adaptability required to generalize to novel tasks and new users.
\IEEEpubidadjcol

To address these limitations, recent studies aim to develop methods with improved adaptability across users and tasks \cite{ref12-1}. For lower-limb exoskeletons, \cite{ref13} designed a human joint motion predictor based on recurrent neural networks, which was validated through simulations in Mujoco \cite{ref14}. \cite{ref15} proposed a human–robot interaction learning and adaptation framework that utilizes DMP and Bayesian optimization to generate personalized walking trajectories online, while employing a task translator to adapt to different tasks. \cite{ref16} and \cite{ref17} developed a task-agnostic controller that significantly reduced user energy expenditure across 28 different tasks. \cite{ref18} employed a data-driven kinematic model to estimate personalized motion features and task states during human walking, enabling real-time torque assistance adjustment. To further enhance individual adaptability, several studies have employed human-in-the-loop (HIL) optimization approaches \cite{ref19}, enabling exoskeletons to provide personalized assistance based on users’ movement preferences during operation and reduce metabolic cost \cite{ref20}, \cite{ref21}. Furthermore, \cite{ref22} leveraged multi-agent reinforcement learning in simulation to generate desired exoskeleton trajectories for a variety of walking tasks. Despite the significant progress in improving user and task adaptability for lower-limb exoskeletons, a notable gap remains in the generalization capabilities of upper-limb exoskeletons, due to the greater flexibility and task variability of upper-limb movements compared to those of the lower limbs.

\begin{figure*}[t]
	\centering
	\includegraphics[width=\textwidth, trim=1 1 1 1, clip]{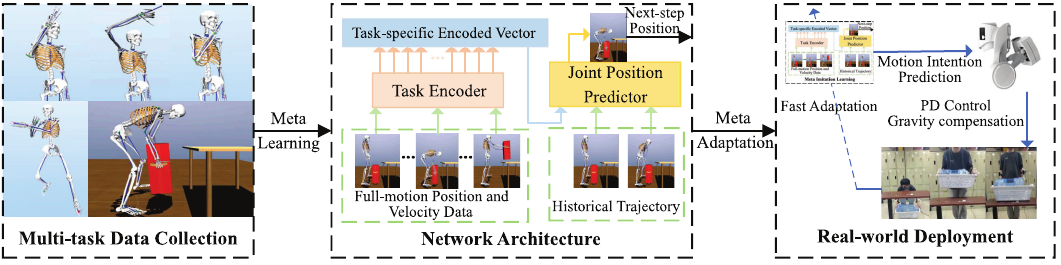}
	\caption{Overall framework of the proposed approach.}
	\label{fig_1}
\end{figure*}

In this article, a novel meta-imitation approach is presented to assist elbow joint movements for different users across a variety of tasks. A task-specific neural network model based on the MAML \cite{ref23} framework is developed to enable exoskeletons to deliver assistance. It consists of a task encoding module and a joint trajectory prediction module constructed using dilated convolutional neural networks. To adapt to new users and tasks, the MAML framework learns an initial set of parameters for the task-specific neural network model that can be quickly fine-tuned to new environments with a few gradient updates. Moreover, to efficiently and accurately acquire diverse motion data, full-body keypoint motions are extracted from motion capture datasets \cite{ref24} and videos \cite{ref25}, covering 42 distinct tasks across three major categories: gesture execution, Tai Chi performance, and object handling. Subsequently, a human skeletal model is constructed within the simulation environment \cite{ref26} to enable retargeting of full-body keypoint motions. Position and velocity data of the elbow joint during flexion are collected to perform meta-learning on the task-specific neural network model. The trained model is directly deployed on an elbow exoskeleton and adapted to new environments through five gradient descent updates. To track the reference trajectories generated by the adapted model, a PD controller with gravity compensation is designed, ensuring accurate and stable motion assistance. The proposed meta-imitation learning framework, along with the data collection and real-world adaptation processes, is illustrated in Fig. \ref{fig_1}.  The contributions of
this article are as follows.

\begin{enumerate}
	\item{A rapid data acquisition pipeline for human joint motion is designed. Full-body keypoints are extracted from publicly available markerless RGB video datasets and marker-based motion capture datasets. A realistic human skeletal model is then constructed in simulation to retarget the extracted keypoint motions, thereby enabling the collection of elbow flexion data aligned with the direction of exoskeleton assistance.
	}
	\item{A task-specific neural network is designed to predict human elbow joint movements. It consists of a task encoder and a trajectory predictor built with dilated convolutional neural networks. The task encoder compresses the entire task trajectory into a low-dimensional vector, which is concatenated with historical position and velocity data and then fed into the predictor to output the next-step elbow joint trajectory.}
	\item{A MAML-based meta-imitation learning approach is applied to train the network. MAML performs inner-loop updates to adapt the model to specific tasks using a few gradient steps, while the outer loop optimizes the model’s initial parameters across a distribution of tasks. The adapted model generates personalized, task-specific trajectories tracked by a PD controller with gravity compensation for stable assistance.}
\end{enumerate}

The rest of this paper is organized as follows. Section II introduces the data acquisition pipeline. Section III presents the joint trajectory prediction method, its training and inference procedures, and the design of the exoskeleton controller. The proposed approach is validated in Section IV through a series of both trained and novel tasks performed by five different users in the real world. Finally, Section V concludes this article.

\section{Method for Elbow Joint Trajectory Data Collection}
\begin{figure}[H]
	\centering
	\includegraphics[width=\columnwidth]{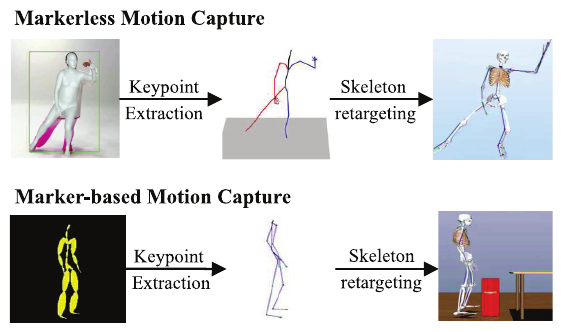}
	\caption{Data collection and preprocessing pipeline for marker-based motion capture data and markerless motion videos.}
	\label{fig_2}
\end{figure}

In this section, The rapid data acquisition pipeline for human joint motion is introduced. The overall workflow is illustrated in Fig. \ref{fig_2}. For both marker-based motion capture data and markerless motion videos, keypoints are extracted and skeletal retargeting is performed.

For markerless motion videos, a Vision Transformer (ViT) is employed to perform both 2D and 3D full-body keypoint estimation \cite{ref25}, and the results are subsequently converted into the skinned multi-person linear (SMPL) format \cite{ref27}. The marker-based motion capture data  is likewise converted into the SMPL format to ensure a unified representation.

To unify human motion data from different sources onto the same skeletal model within Mujoco, skeletal retargeting is performed on the SMPL-format keypoint data. First, a kinematic tree \( \mathcal{T} = (V, E) \) is modeled as a directed acyclic graph representing the skeletal hierarchy, where the vertex set \( V = \{1, 2, \ldots, N\} \) corresponds to \( N \) keypoints, and the edge set \( E \subseteq V \times V \) encodes parent-child relationships. Each edge \( (j, i) \in E \) indicates that keypoint \( j \) is the parent of keypoint \( i \), establishing a hierarchical structure typically rooted at the pelvis or torso. Both the source and target skeletons share the same \( \mathcal{T} \). Given the source skeleton’s 3D keypoint positions \( \{{P}_i^{(s)}\}_{i=1}^N \) defined in the source coordinate frame, the target keypoint positions \( \{{P}_i^{(t)}\}_{i=1}^N \) defined in the target coordinate frame are computed via forward kinematics by recursively propagating local keypoint rotations \( {R}_i^{(t)} \in \mathrm{SO}(3) \) along \( \mathcal{T} \), from the root to the leaves. \( {R}_i^{(t)} \) is obtained by transforming the source rotation \( {R}_i^{(s)} \) via a coordinate transformation as
\begin{equation}\label{eq1}
{R}_i^{(t)} = {Q} \, {R}_i^{(s)} \, {Q}^\top,
\end{equation}
where \( {Q} \in \mathrm{SO}(3) \) represents the rotation from the source coordinate frame to the target coordinate frame.

To compute \( {R}_i^{(s)} \), the unit bone vector \({b}_i^{(s)}\) for each non-root keypoint \(i\) is first defined as
\begin{equation}\label{eq2}
	{b}_i^{(s)} = \frac{{P}_i^{(s)} - {P}_{\mathrm{parent}(i)}^{(s)}}{\left\|{P}_i^{(s)} - {P}_{\mathrm{parent}(i)}^{(s)}\right\|},
\end{equation}
where \( {P}_{\mathrm{parent}(i)}^{(s)} \) is the 3D position of the parent of \( i \) in the source coordinate frame. \( {R}_i^{(s)} \) is then computed to satisfy the following equation
\begin{equation}\label{eq3}
	{R}_i^{(s)} {v}_i^{(s)} = {b}_i^{(s)},
\end{equation}
where \( {v}_i^{(s)} \) denotes the rest pose unit bone vector, measured in the canonical skeleton and expressed in the source coordinate frame, pointing from the parent of \( i \) to \( i \). Based on Rodrigues' rotation formula, \( {R}_i^{(s)} \) has a closed-form expression that defines the rotation aligning \( {v}_i^{(s)} \) with \( {b}_i^{(s)} \). Using \eqref{eq1}, \( {R}_i^{(t)} \) is then obtained.

Finally, with \( {R}_i^{(t)} \) and the root keypoint position \( {P}_{\mathrm{root}}^{(t)} \), each non-root keypoint \( {P}_{i}^{(t)} \) is recursively computed by forward kinematics as
\begin{equation}\label{eq4}
	{P}_i^{(t)} = {P}_{\mathrm{parent}(i)}^{(t)} + {R}_{i}^{(t)}  l_i {v}_i^{(t)},
\end{equation}
where \( {P}_{\mathrm{parent}(i)}^{(t)} \) is the 3D position of the parent of \( i \) in the target coordinate frame, \( {v}_i^{(t)} \) is the rest pose unit bone vector expressed in the target coordinate frame, and \( l_i \) is the bone length measured in the target skeleton’s rest pose. 

Once \( P_i^{(t)} \) are obtained, they are provided to Mujoco’s motion capture interface as reference targets. Mujoco then estimates the generalized human joint configuration \( q_h \in \mathbb{R}^H\) by solving the following inverse kinematics problem
\begin{equation}
	q_h^* = \arg\min_{q_h} \sum_{i=1}^{N} \left\| f_i(q_h) - P_i^{(t)} \right\|^2,
\end{equation}
where \( f_i(q_h) \) denotes the forward kinematics prediction of the \( i \)-th keypoint given \( q_h \), and \( H \) is the number of degrees of freedom in the human model. 

To prepare the dataset for the task-specific network detailed in Section III, the elbow flexion component \( q_h^e \) and its corresponding velocity \( \dot{q}_h^e \) are extracted from \( q_h^* \), where the superscript \( e \) represents the elbow flexion degree of freedom. Each sample consists of a time series of elbow joint angles and angular velocities over the entire task execution, represented as
\begin{equation}
\xi = \left\{ \left( q_{h_k}^e, \dot{q}_{h_k}^e \right) \right\}_{k=1}^{L},
\end{equation}
where \( L \) is the length of the trajectory, \( q_{h_k}^e \) denotes the elbow flexion angle at time step \( k \), and \( \dot{q}_{h_k}^e \) denotes its corresponding angular velocity. The full dataset \( \mathcal{D} \) is  then constructed as the union of datasets corresponding to individual tasks \( \mathcal{D}_m \):
\begin{align}
	\mathcal{D} 
	&= \bigcup_{m=1}^{M} \mathcal{D}_m \nonumber \\
	&= \bigcup_{m=1}^{M} \left\{ \xi_{m,j} \mid j = 1, \dots, N_m \right\},
\end{align}
where \(M\) is the total number of tasks, \(N_m\) is the number of trajectories associated with the task \(m\), and \(\xi_{m,j}\) represents the \(j\)-th trajectory within the task \(m\). 
\section{Task-Specific Network Design, Meta-Learning, and Control Implementation}
In this section, a task-specific neural network is developed to predict human elbow flexion motion, providing personalized reference trajectories for exoskeleton control in movement assistance. The network is trained using MAML to obtain a meta-learned initialization, enabling rapid adaptation to new users and tasks. Finally, the personalized reference trajectories are used in a low-level PD controller with gravity compensation to ensure stable and accurate exoskeleton tracking performance during deployment.
\subsection{Network Architecture}
\begin{figure*}[t]
	\centering
	\includegraphics[width=\textwidth, trim=1 1 1 1, clip]{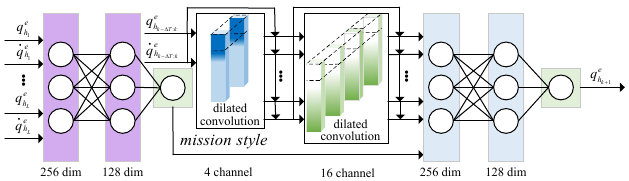}
	\caption{Structure of the task-specific network with task encoding and dilated convolutional layers.}
	\label{fig_3}
\end{figure*}
The architecture of the task-specific neural network is illustrated in Fig. \ref{fig_3}. It consists of a task encoding module and a trajectory prediction module, where the latter employs dilated convolutions for temporal feature extraction. 

The task encoder compresses \( \xi \) into a 128-dimensional latent vector \( z_{\text{task}} \) following an independent normal distribution, which captures the underlying motion characteristics. To enforce this distributional constraint, a Kullback–Leibler (KL) divergence loss 	\( \mathcal{L}_{\text{KL}}\) is applied as follows
\begin{equation}
	\mathcal{L}_{\text{KL}} = D_{\mathrm{KL}} \big( \mathcal{N}(\mu(\xi), \sigma^2(\xi)) \,\|\, \mathcal{N}(0, I) \big).
\end{equation}
To incorporate temporal motion dynamics, at each time step \( k \), a temporal input \( x_k \in \mathbb{R}^{(\Delta T + 1) \times 2} \) is constructed as
\begin{equation}
	x_k = \left\{ \left( q_{h_{k - t}}^e, \dot{q}_{h_{k - t}}^e \right) \right\}_{t = 0}^{\Delta T},
\end{equation}
representing a sequence of elbow flexion angles and angular velocities over the past \( \Delta T + 1 \) steps.
 \( x_k \) is then processed by two layers of one-dimensional dilated convolution with residual connections to capture multi-scale temporal dependencies. The output feature map is subsequently flattened to form a temporal feature vector \( f_k \), which encodes motion dynamics within the time window. \( f_k \) is concatenated with \( z_{\text{task}} \) to form the combined vector. This vector is passed through a feedforward neural network to predict the elbow flexion angle \( q_{h_{k+1}}^e\) at the next time step \( k+1 \). A reconstruction loss \( \mathcal{L}_{\text{rec}} \) measuring the mean squared error between the predicted elbow flexion angle \( \hat{q}_{h_{k+1}}^e \) and \( q_{h_{k+1}}^e \) is expressed as
 \begin{equation}
 	\mathcal{L}_{\text{rec}} = \frac{1}{L} \sum_{k=1}^L \left( \hat{q}_{h_k}^e - q_{h_k}^e \right)^2.
 \end{equation}
The total loss \( \mathcal{L} \) for training combines \( \mathcal{L}_{\text{rec}} \) and \( \mathcal{L}_{\text{KL}} \) is given by
\begin{equation}
	\mathcal{L} = \mathcal{L}_{\text{rec}} + \beta \mathcal{L}_{\text{KL}},
\end{equation}
where \( \beta \) is a weighting factor that balances the two loss terms.
\subsection{Training Method}
To enable fast adaptation to new tasks, MAML is employed to optimize the initial parameters \( \theta \) of the network. For each task \( m \), the corresponding dataset \( \mathcal{D}_m \) is split into two disjoint subsets: a support set \( \mathcal{D}_m^{\text{support}} \) and a query set \( \mathcal{D}_m^{\text{query}} \), such that
\begin{equation}
	\mathcal{D}_m = \mathcal{D}_m^{\text{support}} \cup \mathcal{D}_m^{\text{query}}, \quad \mathcal{D}_m^{\text{support}} \cap \mathcal{D}_m^{\text{query}} = \emptyset.
\end{equation}
During training, \( \theta \) are adapted to each task \( m \) by performing one gradient descent step on the support set loss. The updated parameters \( \theta_m' \) are derived as
\begin{equation}\label{eq13}
	\theta_m' = \theta - \alpha \nabla_{\theta} \mathcal{L}_{\mathcal{T}_m}^{\text{support}}(\theta),
\end{equation}
where \( \alpha \) is the inner-loop learning rate, and \( \mathcal{L}_{\mathcal{T}_m}^{\text{support}}(\theta) \) is the loss computed on the support set \( \mathcal{D}_m^{\text{support}} \).
The adapted parameters \( \theta_m' \) are then evaluated on the query set to compute the meta loss \( \mathcal{L}_{\mathcal{T}_m}^{\text{query}}(\theta_m') \).
Finally, the initial parameters \( \theta \) are updated by optimizing the meta-objective across all tasks
\begin{equation}\label{eq14}
	\theta \leftarrow \theta - \gamma \nabla_{\theta} \sum_{m=1}^M \mathcal{L}_{\mathcal{T}_m}^{\text{query}}(\theta_m'),
\end{equation}
where \( \gamma \) is the meta-learning rate.
By iteratively performing the inner-loop adaptation step \eqref{eq13} and the meta-update step \eqref{eq14}, \( \theta \) are progressively optimized to yield a meta-learned initialization \( \theta^\star \). This initialization enables rapid adaptation to new tasks using only a few gradient steps on limited task-specific data.
\subsection{Online Adaptation}
To accommodate variations in individual users and real-world conditions, an online adaptation phase is introduced. Prior to actuation, the exoskeleton system records the elbow joint motor angle and angular velocity while the user performs the desired task without any motor assistance. The meta-learned initialization \( \theta^\star \) is adapted to the specific user and task by performing a few gradient descent steps using \eqref{eq13}. The resulting parameters \( \theta_{\text{new}}' \) are subsequently used to generate the predicted joint angle \( \hat{q}_h^e \) for the specific task, which serves as the reference command \( q_d \) for the low-level exoskeleton controller. This ensures that the assistance policy is personalized and task-specific.

The dynamics of the single-joint exoskeleton system can be described as
\begin{equation}\label{eq15}
	m(q_r)\ddot{q}_r + c(q_r, \dot{q}_r)\dot{q}_r + g(q_r) = \tau_r,
\end{equation}
where \( q_r \in \mathbb{R} \) and \( \dot{q}_r \in \mathbb{R} \) denote the elbow joint position and velocity of the exoskeleton, respectively. \( m(q_r) \) is the scalar inertia term, \( c(q_r, \dot{q}_r) \) is the Coriolis/centrifugal coefficient, \( g(q_r) \) is the gravity torque, and \( \tau_r \) is the motor input torque. To ensure tracking of the personalized reference command \( q_d \), a PD controller with gravity compensation is applied as
\begin{equation}\label{eq16}
	\tau_r = \hat{g}(q_r) + K_p e + K_d \dot{e},
\end{equation}
where \( e = q_d - q_r \) and \( \dot{e} = - \dot{q}_r \) denote the position and velocity tracking errors, \( K_p > 0 \), \( K_d > 0 \) are control gains, and \( \hat{g}(q_r) \) is the estimated gravity torque. Substituting the control input \eqref{eq16} into the system dynamics \eqref{eq15}, the error dynamics can be expressed as
\begin{equation}\label{eq17}
	m(q_r) \ddot{e} + c(q_r, \dot{q}_r) \dot{e} + K_d \dot{e} + K_p e = \Delta g(q_r),
\end{equation}
where \(\Delta g(q_r) = g(q_r) - \hat{g}(q_r)\) is the gravity compensation error. To analyze the stability of the closed-loop system \eqref{eq17}, consider the following Lyapunov candidate function
\begin{equation}\label{eq18}
	V = \frac{1}{2} m(q_r) \dot{e}^2 + \frac{1}{2} K_p e^2.
\end{equation}
Taking the derivative of \eqref{eq18} and substituting it into \eqref{eq17}, \(\dot{V}\) can be obtained as
\begin{align}\label{eq19}
	\dot{V} &= \frac{1}{2} \dot{m}(q_r) \dot{e}^2 + m(q_r) \dot{e} \ddot{e} + K_p e \dot{e} \nonumber \\
	&= \frac{1}{2}(\dot{m}(q_r) - 2c(q_r, \dot{q}_r)) \dot{e}^2 - K_d\dot{e}^2 + \Delta g(q_r)\dot{e}.
\end{align}
From the skew-symmetry property of the Coriolis term in single degree-of-freedom mechanical systems, it holds that
\begin{equation}\label{eq20}
	\dot{m}(q_r) - 2 c(q_r, \dot{q}_r) = 0.
\end{equation}
According to \eqref{eq20}, \eqref{eq19} can be simplified as
\begin{equation}\label{eq21}
	\dot{V} = - K_d \dot{e}^2 + \Delta g(q_r) \dot{e}.
\end{equation}
By LaSalle's invariance principle, when the gravity compensation error \(\Delta g(q_r) = 0\), \eqref{eq21} satisfies \(\dot{V} \leq 0\). The condition \(\dot{V} = 0\) holds only when \(e = 0\) and \(\dot{e} = 0\). Therefore, all system trajectories converge to the equilibrium point \(e = 0\), guaranteeing asymptotic stability of the closed-loop system \eqref{eq17}.

\section{Experiment design}
\begin{figure}[H]
	\centering
	\subfloat[]{\includegraphics[width=0.48\columnwidth]{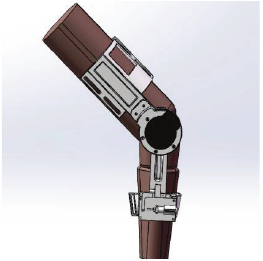}%
		\label{fig_4a}}
	\hfill
	\subfloat[]{\includegraphics[width=0.48\columnwidth]{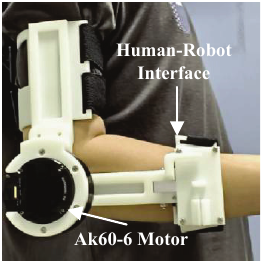}%
		\label{fig_4b}}
	\caption{Mechanical design and physical prototype of the exoskeleton. (a) SolidWorks model. (b) Real-world elbow exoskeleton with 3D-printed components and joint motor.}
	\label{fig_4}
\end{figure}

In this study, experiments are conducted on an elbow joint exoskeleton developed by \cite{ref28}, as depicted in Fig. \ref{fig_4}. The design features direct alignment between the motor shaft and the human elbow joint, and employs a CubeMars AK60-6 motor for actuation. The exoskeleton also integrates a 3D-printed joint limit component to enhance user safety and restrict motion within a safe range. The effectiveness of the exoskeleton assistance is assessed under cross-task and cross-user scenarios by quantifying muscle activation and respiratory metabolic expenditure. All experiment procedures
have been reviewed and approved by the ethics committee of
Institute of Automation, Chinese Academy of Sciences under
the protocol no. IA21-2302-140202.

\subsection{Cross-Task Performance Analysis}
To evaluate the task generalization capability of the proposed method, the exoskeleton is tested on five diverse tasks involving both unilateral and bilateral upper-limb movements. These tasks include:
\begin{enumerate}
	\item Performing the OK gesture with one hand
	\item Carrying a 6\,kg pack of bottled water with both hands
	\item Practicing Tai Chi Form 10 with both hands
	\item Lifting the dumbbell with one hand
	\item Lifting a 20\,kg water bucket with both hands
\end{enumerate}
Tasks 1, 3, and 4 involve task trajectories that are not included in the training data used for meta-learning. Although Tasks 2 and 5 involve motions similar to those in the training data, the specific objects carried are novel. To adapt to novel tasks and users, each user first performs a demonstration of the assigned task while wearing the exoskeleton without assistance. The exoskeleton records the motor motion data throughout the task, which is subsequently used to update the meta-trained model. After five gradient descent steps, an adapted network policy is obtained for the specific task and user.

The gravity torque term in controller \eqref{eq16} is designed as
\begin{equation}\label{eq22}
	\hat{g}(q_r) =  mgl_m\sin(q_r),
\end{equation}
where \( m \) is the estimated load mass, \( g \) is gravitational acceleration, and \( l_m \) is the moment arm. \eqref{eq22} compensates for the gravity of the dumbbell or the carried object.

\begin{figure*}[t]
	\centering
	\includegraphics[width=\textwidth, trim=1 1 1 1, clip]{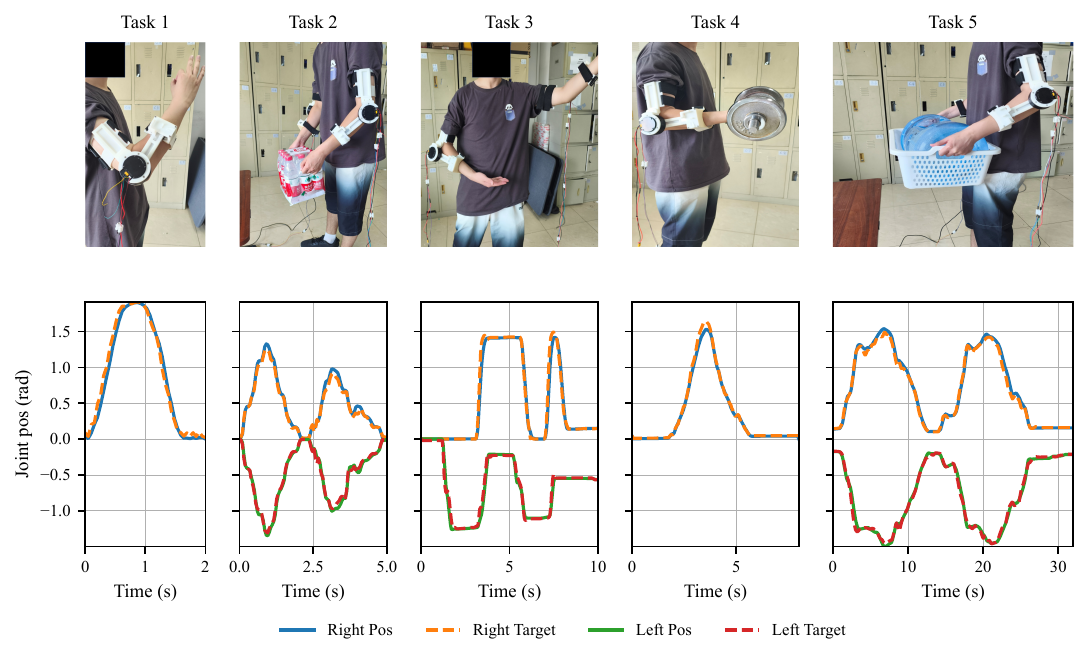}
	\caption{Task illustration photos and joint trajectory comparisons for five tasks. The top row shows illustrative photos representing each task. The bottom row presents the corresponding joint positions over time, where solid lines indicate the average actual positions and dashed lines indicate the average target positions, both computed over five repeated trials.}
	\label{fig_5}
\end{figure*}

\begin{figure}[h]
	\centering
	\includegraphics[width=\columnwidth]{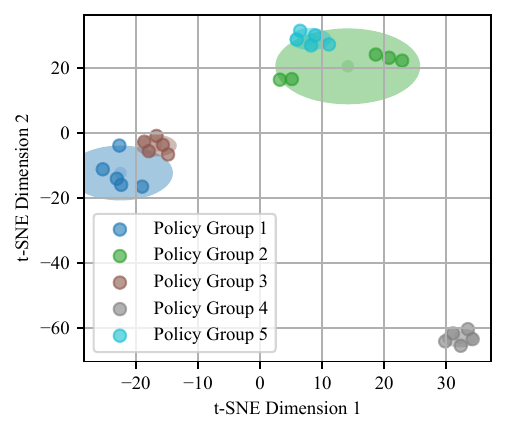}
	\caption{Visualization of the policy feature distributions learned for the five tasks at the right elbow joint. Each solid dot represents a sampled policy feature, with different colors corresponding to different task-specific policies.}
	\label{fig_6}
\end{figure}

Figure \ref{fig_5} shows the average desired and actual joint trajectories of the exoskeleton over five repeated trials across the five different tasks introduced above, along with representative photos illustrating the task scenarios. A complete video demonstrating the execution of all tasks is provided in the supplementary material. The desired joint trajectories are generated by five task-specific adapted neural networks, while the actual joint trajectories are obtained by driving the exoskeleton through the controller \eqref{eq16}. The average root-mean-square (RMS) tracking error across the five tasks is 0.056 rad. The maximum tracking error is 0.093 rad, observed in Task 1, while the minimum tracking error is 0.038 rad, observed in Task 4. To illustrate the differences among the five task-specific adapted neural network policies, t-distributed stochastic neighbor embedding (t-SNE) is employed to visualize their feature distributions at the right elbow joint, as shown in Fig. \ref{fig_6}. Notably, the policies for Task 2 and Task 5 exhibit similar feature patterns, which can be attributed to the similarity of their task trajectories.

During task execution with exoskeleton assistance, the surface electromyography (sEMG) signals of the Brachioradialis (BRR) muscle are measured to evaluate the assistance effectiveness. Compared with performing the tasks without wearing the exoskeleton, the average reductions in muscle activation over five repeated trials are summarized in Table \ref{Table_1}. Furthermore, in order to assess the overall level of assistance provided by the exoskeleton, respiratory metabolic data are collected using a metabolic mask during Task 5, the most physically demanding among all tasks. Compared to the unassisted condition, the calorie expenditure decreases by an average of 29.31\% with exoskeleton assistance.
\begin{table}[H]
	\caption{Average sEMG Reduction (\%) for Left and Right Arms Across Tasks}
	\label{Table_1}
	\centering
	\begin{tabular}{c c c c c c}
		\toprule
		Task ID & 1 & 2 & 3 & 4 & 5 \\
		\midrule
		Left Arm (\%) & -- & 35.77 & 35.66 & -- & 34.92 \\
		Right Arm (\%) & 55.53 & 28.11 & 55.79 & 12.94 & 20.45 \\
		\bottomrule
	\end{tabular}
\end{table}
\subsection{Cross-User Performance Analysis}
Five users with an average age of 26 years are selected for this experiment. Their heights range from 170 to 185 cm and their weights range from 50 to 80 kg. None of the participants have prior experience using the exoskeleton. Each subject is instructed to perform Tasks 1, 2, and 4 under two conditions: (1) with exoskeleton assistance and (2) during natural, unassisted motion. 

\begin{figure*}[t]
	\centering
	\includegraphics[width=\textwidth, trim=1 1 1 1, clip]{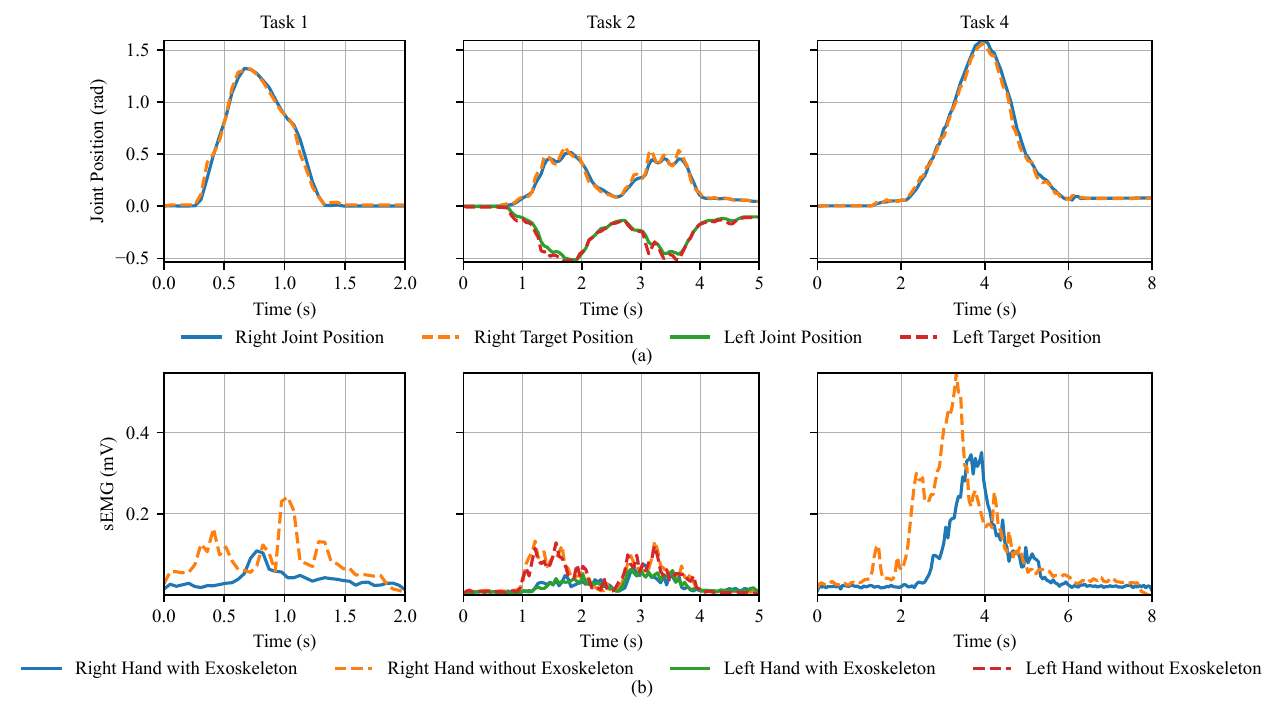}
	\caption{Average joint trajectory of the exoskeleton and average sEMG changes of five subjects during Task 1, 2, and 4. (a) Comparison between the average desired trajectory and the actual trajectory of the exoskeleton with assistance. (b) Average sEMG signal changes of the subjects under two conditions: with exoskeleton assistance and natural, unassisted motion.}
	\label{fig_7}
\end{figure*}

Figure~\ref{fig_7} presents the average desired and actual joint trajectories of the exoskeleton during three distinct tasks performed by five subjects, along with the corresponding average sEMG amplitude changes under both assisted and unassisted conditions. The maximum RMS tracking error is 0.042~rad, observed in Task~1, while the minimum is 0.028~rad in Task~2. The results demonstrate a consistently high level of control accuracy across the evaluated tasks. A reduction in sEMG amplitude is observed across all tasks when the exoskeleton provides assistance, indicating decreased muscle activation. The largest reduction occurs in Task~1, with an average decrease of 54.16\%, while the smallest reduction is observed in Task~4, at 37.38\%. These findings suggest that the exoskeleton effectively reduces users’ muscular effort during task execution.

\begin{figure*}[t]
	\centering
	\includegraphics[width=\textwidth, trim=1 1 1 1, clip]{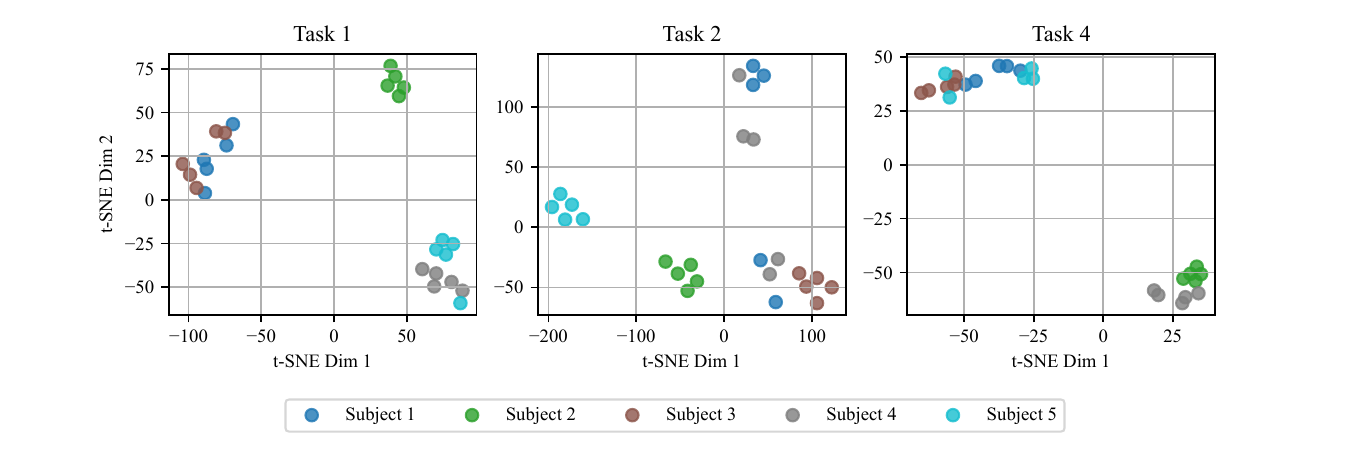}
	\caption{t-SNE visualization of policy features learned from five subjects at the right elbow joint. Each dot represents a sampled feature from one subject's policy, with different colors indicating different subject-specific strategies.}
	\label{fig_8}
\end{figure*}
To further investigate the adaptability of network policies across different subjects, t-SNE is used to visualize the policy feature distributions for each task, revealing the structural properties and separability of the learned representations among subjects. Fig. \ref{fig_8} illustrates the differences in policy feature distributions among the five subjects after adaptation to the same task. The results indicate that, in most cases, learned policies differ significantly across subjects when performing the same task, whereas policies learned by the same subject at different times show relatively minor variations. This  is likely due to considerable variability in trajectory-related factors such as amplitude, angular velocity, and task completion time. However, when different subjects perform the same task with similar trajectories, their learned policies tend to be similar as well. For example, in Task 1, the policies of Subject 1 and Subject 3 are close to each other, as are those of Subject 4 and Subject 5. In Task 4, the policies of Subjects 1, 3, and 5 form one cluster, while those of Subjects 2 and 4 form another. For Task 2, learned policies can distinguish Subject 2 and Subject 5, whereas the other three subjects cannot be clearly differentiated. This may be due to an insufficient sample size.

\section{Conclusion}
This paper proposes a meta-imitation learning framework for wearable exoskeletons that enables personalized and task-generalizable assistance. A fast data acquisition pipeline is developed by retargeting human motion from video and motion-capture datasets. Leveraging the acquired data, a task-specific neural network is trained within the MAML framework to facilitate rapid adaptation to novel users and tasks.  After adaptation in real-world environments, the trained network generates personalized, task-specific trajectories tracked by a low-level controller to effectively assist users. Experimental results demonstrate that the proposed method enables the exoskeleton to provide effective assistance to diverse users across a range of real-world tasks.


%
%
%
%
%
%
%

\end{document}